\documentclass[11pt]{article}

\usepackage[preprint]{acl}

\usepackage{times}
\usepackage{latexsym}

\usepackage[T1]{fontenc}
\usepackage{amsthm}

\newtheorem{assumption}{Assumption}
\usepackage[utf8]{inputenc}

\usepackage{microtype}

\usepackage{inconsolata}

\usepackage{graphicx}

\usepackage{graphicx}
\usepackage{bm}
\usepackage{amsmath}
\usepackage{amssymb}
\usepackage{amsfonts}
\usepackage{mathtools}
\usepackage{bbm}
\usepackage{multirow}
\usepackage{multicol}
\usepackage{booktabs}
\usepackage{colortbl}
\usepackage{footmisc}
\usepackage{fontawesome}
\usepackage{epigraph}
\newcommand{\thicktilde}[1]{\mathbf{\tilde{\text{$#1$}}}}
\newcommand{\shortName}[1]{\textsc{FrugalPrompt}}

\title{\shortName\ : Reducing Contextual Overhead in Large\\Language Models via Token Attribution}

\author{
  Syed Rifat Raiyan\textsuperscript{$1$,*}, Md Farhan Ishmam\textsuperscript{$2$,*}, \\
  \textbf{Abdullah Al Imran}\textsuperscript{$3$}, \textbf{Mohammad Ali Moni}\textsuperscript{$4$}\vspace{0.2cm}\\
  \textsuperscript{$1$}Islamic University of Technology, Bangladesh \quad \textsuperscript{$2$}University of Utah, USA \\ \textsuperscript{$3$}University of Liverpool, UK \quad \textsuperscript{$4$}University of Queensland, Australia \\
  \texttt{rifatraiyan@iut-dhaka.edu, farhan.ishmam@utah.edu}\quad $^*$Equal Contribution\vspace{0.3cm} \\ 
  \small \faGithub\;\href{https://github.com/Starscream-11813/Frugal-ICL}{\texttt{Starscream-11813/Frugal-ICL}} \qquad \faGlobe\;\href{https://frugalprompt.github.io/}{\texttt{frugalprompt.github.io}}
}
\newtheorem{theorem}{Theorem}
\newtheorem{corollary}{Corollary}

\begin{document}
\maketitle
\begin{abstract}
Human communication heavily relies on laconism and inferential pragmatics, allowing listeners to successfully reconstruct rich meaning from sparse, telegraphic speech. In contrast, large language models (LLMs) owe much of their stellar performance to expansive input contexts, yet such verbosity inflates monetary costs, carbon footprint, and inference-time latency. This overhead manifests from the redundant low-utility tokens present in typical prompts, as only a fraction of tokens typically carries the majority of the semantic weight. Inspired by the aforementioned cognitive psycholinguistic processes, we address this inefficiency by introducing \textsc{FrugalPrompt}, a novel prompt compression framework for LLMs, which retains only the most semantically significant tokens. Leveraging two state-of-the-art token attribution methods, GlobEnc and DecompX, we assign salience scores to every token in an input sequence, rank them to retain the top-$k\%$ tokens, and obtain a sparse \textit{frugalized} prompt. 
We establish the theoretical stability of our approach and provide strong empirical results across a suite of four NLP tasks to study the trade-off between the portion of retained tokens and performance. Experimental findings across retention settings reveal asymmetric performance patterns that suggest potential task contamination effects. 
We posit that our work contributes to a more nuanced understanding of LLM behavior in performance-efficiency trade-offs and delineates the boundary between tasks tolerant of contextual sparsity and those requiring exhaustive context. 



\end{abstract}
\vspace{-3mm}
\epigraph{\itshape ``Why waste time say lot word when few word do trick?''}{---Kevin Malone, \textit{The Office} (Season 8, Episode 2)}
\vspace{-3mm}
\section{Introduction}

As large language models (LLMs) are now becoming general-purpose intelligence systems, they are supporting workflows that go beyond their initial zero-shot text generation: from coding systems capable of using tools to multi-step agentic systems solving multimodal problems \cite{yang2024swe, schmidgall2025agent}. Yet the same empirical scaling laws that fuel these capabilities into trillion-parameter models \cite{fedus2022switch, hudson2023trillion} also intensify a basic cognitive and computational tension. Human language is richly \textit{redundant}\footnote{We define redundancy as any verbiage that extends beyond the most succinct form of communication.}, while modern LLM systems being trained and conditioned on human text are constrained by context length \cite{wu2025inference}, latency \cite{wan2023efficient}, and energy costs \cite{stojkovic2024towards} due to the {verbosity} of human language.



A core observation of studying linguistics from the lens of information theory is that natural language is not {an} information-minimal code \cite{shannon1951prediction}. Human utterances inherently contain statistical redundancy, \textit{e.g.}, text is repeated for clarification, and function words {serve grammatical rather than semantic roles}. Many tokens, therefore, contribute little incremental information once the local context is known. Psycholinguistic evidence similarly suggests that comprehenders can often recover intended meaning from “telegraphic” input, a property exploited in simplified registers and second-language communication \cite{gibson2013rational,klein1997basic}. These findings suggest a cognitively motivated route to efficiency in NLP, and we ask: To what extent do LLMs require full linguistic explicitness, and to what extent can they reconstruct meaning from pragmatically sufficient cues?

We aim to answer these questions with a theoretically grounded work motivated by information-theoretic approaches in linguistics. Using SOTA token attribution techniques, we compute saliency scores for the prompt and selectively retain the semantically rich tokens while filtering redundant ones. This yields a training-free compressor that (i) better preserves meaning-critical content, (ii) systematically removes low-value discourse material and function words, and (iii) avoids the cost of an auxiliary large model. {Importantly, our approach requires only a lightweight pre-trained encoder, where the saliency scores are generated by state-of-the-art token attribution methods \cite{modarressi-etal-2022-globenc, modarressi-etal-2023-decompx} applied to a single 110M-parameter BERT model \cite{devlin2019bert}, incurring substantially lower parameter overhead than the auxiliary language models used by prior compressors \cite{Jiang2023LLMLinguaCP, Chuang2024LearningTC}.} 

Our filtering criteria are grounded in two linguistic principles underlying textual redundancy. First, there is an intrinsic redundancy to any form of text produced by humans \cite{wit1999linguistic}. For instance, in the text, \textit{``What are your plans for the weekend, by the way?''}, the phrase \textit{``by the way''} contributes no additional meaning to an LLM. Second, the English language has low-importance tokens that add little to no semantic weight, generally in the form of articles and auxiliary verbs. For instance, most of the meaning is conveyed if the statement \textit{``I have received a parcel.''}, is simplified to \textit{``I received parcel.''}, with the latter mimicking simplified English, often used by non-native speakers \cite{shubert1995comprehensibility}. Our filtering methods eliminate both forms of redundancy. 
Our key contributions are as follows:

\begin{itemize}
    \item We introduce a novel, training-free prompt frugalization strategy for LLMs that controllably filters low-importance tokens based on saliency scores from pre-trained encoders. 

    \item Our method is theoretically grounded with empirical results showing strong retention–performance trade-offs. Across four NLP tasks, a $20\%$ prompt reduction preserves performance for most models with negligible parameter overhead.
    
    \item We reveal behavioral insights into how scaling performance relates to inference cost and potential task contamination in benchmarks. 
\end{itemize}





\section{Related Work}
\label{sec:relatedWork}

\subsection{Redundancy in Language}
\label{subsec:relatedWorkLanguageredundancy}
Language, by nature, is not {an} information-minimal code \cite{shannon1951prediction} and introduces several forms of redundancy. Through the lens of information theory \cite{shannon1948mathematical}, redundancy of a language can be defined as the difference between the maximum possible entropy of the language and its actual entropy. Theories in inferential pragmatics \cite{Richards1990InferentialPragmatics} and relevance-driven communication argue that interlocutors reconstruct implicit content by focusing on semantically informative cues rather than processing every token with equal weight. Such laconism is not an anomaly but a fundamental property of human communicative efficiency \cite{LevshinaMoran2021Efficiency}.

\subsection{LLM Efficiency} 
The current literature on LLM efficiency \cite{wan2023efficient} stems from earlier works on efficient transformers \cite{efficientTransformer}. Inference-time efficiency can be achieved via model compression methods, such as quantization \cite{dettmers2023spqr}, parameter pruning \cite{ma2023llmpruner}, low-rank decomposition \cite{Yixiao2023loSparse}, and knowledge distillation \cite{timiryasov-tastet-2023-babyllama}. Training efficiency can be achieved during pre-training \cite{liu2023sophia,yao2024masked} and fine-tuning, either through parameter-efficient \cite{hu2022lora} or memory-efficient \cite{dettmers2023qlora} approaches. Finally, architectural changes also improve efficiency, \textit{e.g.}, efficient attention \cite{Child2019GeneratingLS, ainslie-etal-2023-gqa} and state-space models \cite{Gu2023MambaLS}. 


Distinct from these methods, our work adopts a data-centric approach, where the input prompt is optimized rather than the model's underlying weights or architecture \cite{Chang2024EfficientPM}. In essence, we perform prompt compression, which minimizes input redundancy while retaining task-specific information{ \cite{Li2023CompressingCT, Jiang2023LLMLinguaCP, pan-etal-2024-llmlingua}.} Contrasting previous approaches that often use a paraphrasing model \cite{Chuang2024LearningTC}, which can lead to instability, ours takes a theoretically grounded approach by filtering the least salient tail-end tokens. {Among token-level compressors, Selective Context \cite{Li2023CompressingCT} prunes lexical units with low self-information under a causal language model, while LLMLingua-2 \cite{pan-etal-2024-llmlingua} distills GPT-4 compression annotations into a trained token classifier.}



\subsection{Token Attribution}
The self-attention weights in transformers have been primarily explored as a proxy for token importance \cite{htut2019attention,clark2019does}, yet they fail to align with the true attribution of tokens \cite{jain2019attention}. While including vector norms can partially mitigate this issue \cite{kobayashi-etal-2020-attention}, it overlooks the intricacies of the non-attention components in a transformer. Instead, we rely on subsequent token attribution methods (see \autoref{sec:prelemTokenAttribution}): GlobEnc \cite{modarressi-etal-2022-globenc}, which quantifies global token attribution by aggregating layer-wise analyses across the entire encoder using attention rollout and vector norms, and DecompX \cite{modarressi-etal-2023-decompx}, which propagates locally decomposed token representations globally to incorporate all model components, including non-linear feed-forward networks. These methods were selected for their superior ability to capture comprehensive semantic contributions beyond raw attention weights, enabling identification of salient tokens with minimal computational overhead.





\begin{figure*}
    \centering
    \includegraphics[width=\linewidth]{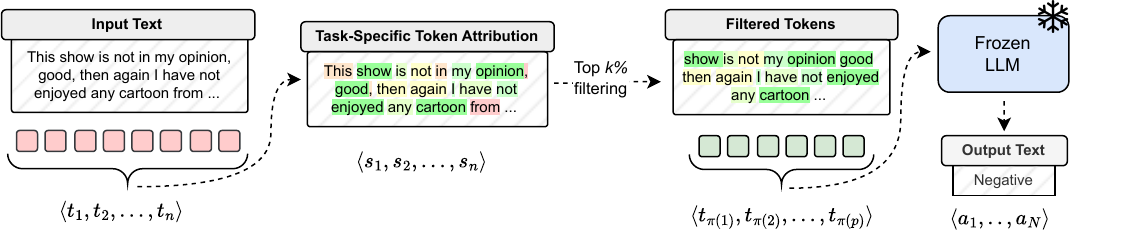} 
    \caption{
    Overview of the \textsc{FrugalPrompt} pipeline. The input prompt with tokens $\langle t_1, t_2,\dots,t_n\rangle$ is passed to our task-specific token attribution module to generate the saliency scores for each token $\langle s_1, s_2,\dots,s_n\rangle$. The tokens are ordered based on their saliency in a permutation $\pi$ and the top {$p=\lceil\frac{k}{100}\cdot n\rceil$ tokens are retained} while maintaining the order of the text to form the reduced text $\langle t_{\pi(1)}, t_{\pi(2)},\dots,t_{\pi(p)}\rangle$. The reduced text is inferred by a frozen LLM to generate the answer tokens {$\langle a_1, a_2,\dots,a_N\rangle$}.
    }
    \label{fig:pipeline}
\end{figure*}


\section{Methodology}
\label{sec:methodology}


Figure \ref{fig:pipeline} portrays an overview of our proposed pipeline. We obtain the saliency scores using the token attribution methods: GlobEnc \cite{modarressi-etal-2022-globenc} and DecompX \cite{modarressi-etal-2023-decompx}, {(details in \S~\ref{sec:prelemTokenAttribution}),} and employ frugalization via token filtration to generate a condensed yet semantically representative version of an input sample. This technique identifies and retains only the most salient tokens, thereby reducing input dimensionality while preserving the core meaning for the given task. {Inputs longer than the encoder's 512-token window are scored via sentence-level chunking, as detailed in Appendix~\ref{app:longInputs}.}
\subsection{Ranking via Saliency Scores}
We define a task-specific scoring function {$\varphi_\tau\in\{\text{GlobEnc},\ \text{DecompX}\}$} for a task $\tau$, which assigns a saliency score to each token in a given text, \textit{i.e.,}
{\setlength{\abovedisplayskip}{3pt}
 \setlength{\belowdisplayskip}{3pt}
\begin{align*}
\varphi_\tau:\mathcal{T}\rightarrow\mathbb{R}^{n},
\end{align*}
}
where $\mathcal{T}$ denotes the space of the token sequences and $n$ is the length of the sequence. For the input token sequence, $T = \langle t_1, t_2, \dots, t_n\rangle$, the scoring function produces a saliency vector,
{\setlength{\abovedisplayskip}{3pt}
\setlength{\belowdisplayskip}{3pt}
\begin{align*}
\mathbf{s} = \varphi_\tau(T) =\langle s_1,s_2,\dots,s_n\rangle,\quad s_i\in\mathbb{R},
\end{align*}
}
where each score $s_i$ reflects the relative contribution of token $t_i$ to the model's output for the specified task.
Tokens are then ranked in monotone decreasing order of their attribution scores to obtain a permutation, \(\pi\), formed of the token indices, such that, 
{\setlength{\abovedisplayskip}{2pt}
\setlength{\belowdisplayskip}{2pt}
\begin{align*}
s_{\pi(1)} \ge s_{\pi(2)} \ge \cdots \ge s_{\pi(n)}.
\end{align*}
}
\subsection{\texorpdfstring{Top-$k$ Filtering}{Top-k Filtering}} 
\label{sec:topkFiltering}
To construct the \textit{frugalized representation}, we retain only the top \(p = \left\lceil \frac{k}{100} \cdot n \right\rceil\) tokens, where $k$ is the retention percentage. The selected indices form the sequence,  
{\setlength{\abovedisplayskip}{2pt}
\setlength{\belowdisplayskip}{2pt}
\begin{align*}
{S}_k = \langle \pi(1), \pi(2), \dots, \pi(p)\rangle.
\end{align*}
}
We reindex $S_k$ in monotone increasing order to produce ${S}_{k\uparrow}$. The frugalized sequence \({F}_k\) is produced by restoring the selected tokens to their original order, \textit{i.e.},
{\setlength{\abovedisplayskip}{2pt}
\setlength{\belowdisplayskip}{2pt}
\begin{align*}
{F}_k = \langle t_i \lvert i \in {S}_{k\uparrow}\rangle
\end{align*}
}
ensuring that the temporal coherence of the sequence is preserved, while uninformative tokens are discarded.

We elucidate our frugalization approach with an example.
For the input sequence \( {T} = \) $\langle$\texttt{``The''}, \texttt{``movie''}, \texttt{``was''}, \texttt{``good''}, \texttt{``,''}, \texttt{``and''}, \texttt{``I''}, \texttt{``liked''}, \texttt{``it''}, \texttt{``very''}, \texttt{``much''}$\rangle$, suppose the attribution scores for sentiment analysis prioritize content and sentiment words (\textit{e.g.}, \texttt{``movie''}: $0.95$, \texttt{``good''}: $0.90$, \texttt{``much''}: $0.85$, \texttt{``liked''}: $0.80$). Upon {retaining the top $k = 35\%$ (\( p = \lceil 0.35\cdot 11\rceil = 4 \))} and reordering them according to their original relative order, we get the reduced sequence
{\setlength{\abovedisplayskip}{2pt}
\setlength{\belowdisplayskip}{2pt}
$${F}_{{35}} = \langle\texttt{``movie''}, \texttt{``good''}, \texttt{``liked''}, \texttt{``much''}\rangle$$}
where the original order is preserved to maintain syntactic coherence. The frugalized output \emph{``movie good liked much''} is thus created, discarding the less salient tokens (\textit{e.g.}, \texttt{``The''}, \texttt{``was''}, etc.) while retaining semantic essence.

\subsection{Theoretical Bounds and Relationships}
\label{sec:theoreticalBounds}
We now formalize when frugalization is safe. Our main result (Corollary~\ref{cor:del-k}) shows that the loss incurred by retaining only the top-$k\%$ salient tokens is bounded by two terms: (i) a linear term in the saliencies of the \emph{deleted} tokens, which is minimized by our top-$k$ rule, and (ii) a second-order interaction term that grows quadratically with the number of deletions. This decomposition explains why top-$k$ filtering works and aggressive deletion eventually breaks down. The bounds formalize the relationship between dropping $(100-k)\%$ of tokens and the estimated performance.
\paragraph{Deletion Effect.} For a filtered or deleted set of indices $D\subseteq[n]=\{1,\dots,n\}$, we define the retained subsequence of tokens $T\setminus D := T_{[n]\setminus D}$. For task $\tau$ and model parameters $\theta$, let $\ell_\tau(T,y;\theta)\in\mathbb{R}^+$ be the task loss or evaluation score. We define the deletion-effect set function:
{\setlength{\abovedisplayskip}{2pt}
\setlength{\belowdisplayskip}{2pt}
\begin{equation}
f(D) \;:=\; \underbrace{\ell_\tau(T\setminus D,y;\theta)}_{\text{Loss after Deletion}} - \underbrace{\ell_\tau(T,y;\theta)}_{\text{Original Loss}}.
\label{eq:del-set-fn}
\end{equation}
}
Trivially, $f(\emptyset)=0$. We emphasize that all results are inference-time, \textit{i.e.}, the model parameters $\theta$ are fixed. We treat $\ell_\tau$ as a loss throughout (lower is better), so that $f(D)>0$ indicates deletion hurts performance; for metrics where higher is better (\textit{e.g.}, accuracy, ROUGE), $\ell_\tau$ should be read as the negative of that metric.
\paragraph{Discrete Marginals and Interactions.}
To bound $f(D)$ when $|D|$ is large, we decompose it into per-token marginals (Eq.~\ref{eq:firstDef}) and pairwise interactions (Eq.~\ref{eq:second-def}), and the assumptions below then control each piece separately.
For a set of deleted tokens $A\subseteq[n]$ and a new token $i\notin A$, we define the marginal deletion:
{\setlength{\abovedisplayskip}{2pt}
\setlength{\belowdisplayskip}{2pt}
\begin{equation}
\Delta_i f(A) \;:=\; f(A\cup\{i\}) - f(A).
\label{eq:firstDef}
\end{equation}
}
For tokens $i\neq j$ with $i,j\notin A$, we define the second-order discrete difference, \textit{i.e.}, the pairwise interaction as
{\setlength{\abovedisplayskip}{2pt}
\setlength{\belowdisplayskip}{2pt}
\begin{align}
  \nonumber  \Delta_{ij} f(A) \;:=\; \Delta_i\Delta_j f(A) = f(A\cup\{i,j\}) \\- f(A\cup\{i\}) - f(A\cup\{j\}) + f(A).
  \label{eq:second-def}
\end{align}
}
\begin{assumption}[Saliency upper-bounds singleton deletion harm]
\label{assump:del}
There exists a task constant, $C_\tau\ge 0$, such that,
{\setlength{\abovedisplayskip}{2pt}
\setlength{\belowdisplayskip}{2pt}
\begin{align}
f(\{i\}) \;&\le\; C_\tau\, s_i,\quad\forall i\in[n].\label{eq:singleton-dom}
\end{align}
}
\end{assumption}
Assumption~\ref{assump:del} is a faithfulness condition: it asserts that the attribution score $s_i$ is not merely a heuristic ranking but a calibrated upper bound (up to the task-dependent constant $C_\tau$) on the harm of removing token~$i$ in isolation. This is precisely the property that GlobEnc and DecompX are designed to provide, and is empirically supported by the faithfulness evaluations in~\citet{modarressi-etal-2022-globenc, modarressi-etal-2023-decompx}. We treat $C_\tau$ as an empirical constant that absorbs any departure from the ideal faithfulness for task~$\tau$.

\begin{assumption}[Bounded pairwise interaction]\label{assump:pairwiseBound} There exists a task constant, $\gamma_\tau\ge0$, such that,
{\setlength{\abovedisplayskip}{2pt}
\setlength{\belowdisplayskip}{2pt}
\begin{align}
\Delta_{ij} f(A) \;&\le\; \gamma_\tau, \quad \forall A\subseteq[n],\forall i,j\notin A.\label{eq:bounded-interaction}
\end{align}
}
\end{assumption}
Assumption~\ref{assump:pairwiseBound} is a submodular-flavored condition standard in coverage and feature-selection theory: it prevents the joint harm of deleting two tokens from exceeding the sum of their individual harms by more than a constant $\gamma_\tau$. We impose only the one-sided bound because negative interactions are benign (and indeed expected when deleted tokens are mutually redundant): two redundant tokens deleted together may hurt less than the sum of their singleton deletions. The constant $\gamma_\tau$ thus captures the worst-case \emph{compositional} fragility of task~$\tau$ to multi-token deletion.



\begin{theorem}[Deletion bound via saliency and interaction]
\label{thm:del-bound}
Under {Assumptions~\ref{assump:del} and~\ref{assump:pairwiseBound}}, for any deletion set $D$ with $|D|=q$,
{\setlength{\abovedisplayskip}{2pt}
\setlength{\belowdisplayskip}{2pt}
\begin{equation}
f(D)
\;\le\;
C_\tau\sum_{i\in D} s_i
\;+\;
\frac{\gamma_\tau}{2}\,q(q-1).
\label{eq:del-bound}
\end{equation}
}
\end{theorem}

\definecolor{mygray}{gray}{0.9}
\definecolor{mycyan}{RGB}{189, 224, 252}

\begin{table*}[h!]
\setlength{\tabcolsep}{3pt}
\renewcommand{\arraystretch}{0.9} 
\centering
\small
\begin{tabular}{@{}lcccccccccccc@{}}
\toprule
\multicolumn{1}{c}{\multirow{3}{*}{\textbf{Model}}} & \multirow{4}{*}{\textbf{\begin{tabular}[c]{@{}c@{}}Prompt\\ Reduction\\ Method\end{tabular}}} & \multirow{3}{*}{\textbf{$k\%$}} & \multicolumn{10}{c}{\textbf{Tasks}} \\ \cmidrule(l){4-13}
\multicolumn{1}{c}{} &  &  & \multicolumn{2}{c}{\textbf{CLS}} & \multicolumn{6}{c}{\textbf{SUM}} & \textbf{QA} & \textbf{RSN} \\ \cmidrule(lr){4-5} \cmidrule(lr){6-11} \cmidrule(lr){12-12} \cmidrule(lr){13-13}
\multicolumn{1}{c}{} &  &  & \textbf{Acc} & \textbf{F1} & \textbf{BLEU} & \textbf{R-1} & \textbf{R-2} & \textbf{R-L} & \textbf{BERT} & \textbf{METEOR} & \textbf{Acc} & \textbf{pass@1} \\ \midrule
 
\multirow{13}{*}{Llama-3 8B} & \multicolumn{1}{c}{N/A} & \multicolumn{1}{c}{100}   & \cellcolor{mygray}0.949 & \cellcolor{mygray}0.949 & \cellcolor{mygray}0.020 & \cellcolor{mygray}0.232 & \cellcolor{mygray}0.073 & \cellcolor{mygray}0.193 & \cellcolor{mygray}0.872 & \cellcolor{mygray}0.335 & \cellcolor{mygray}0.813 & \cellcolor{mygray}0.786 \\ \cmidrule(l){2-13}
 & \multirow{3}{*}{GlobEnc} & 80 & \cellcolor{mycyan}\textbf{0.942} & \cellcolor{mycyan}\textbf{0.942} & \cellcolor{mycyan}\textbf{0.017} & \cellcolor{mycyan}\textbf{0.226} & \cellcolor{mycyan}\textbf{0.069} & \cellcolor{mycyan}\textbf{0.189} & \cellcolor{mycyan}\textbf{0.871} & \cellcolor{mycyan}\textbf{0.330} & \cellcolor{mycyan}\textbf{0.798} & 0.500 \\
 &  & 60 & \cellcolor{mycyan}\textbf{0.942} & \cellcolor{mycyan}\textbf{0.942} & 0.015 & 0.220 & 0.064 & 0.181 & 0.870 & 0.319 & 0.754 & 0.243 \\
 &  & 50 & 0.921 & 0.921 & 0.013 & 0.213 & 0.057 & 0.175 & 0.868 & 0.303 & 0.716 & 0.146 \\ \cmidrule(l){2-13}
 & \multirow{3}{*}{DecompX} & 80 & 0.936 & 0.936 & 0.015 & 0.218 & 0.061 & 0.179 & 0.869 & 0.311 & 0.747 & 0.652 \\
 &  & 60 & 0.900 & 0.900 & 0.014 & 0.206 & 0.055 & 0.169 & 0.866 & 0.286 & 0.681 & 0.425 \\
 &  & 50 & 0.868 & 0.868 & 0.012 & 0.198 & 0.048 & 0.161 & 0.864 & 0.270 & 0.670 & 0.265 \\ \cmidrule(l){2-13}
 & \multirow{3}{*}{LLMLingua2} & 80 & 0.920 & 0.919 & 0.014 & 0.225 & 0.068 & 0.185 & 0.864 & 0.323 & 0.790 & \cellcolor{mycyan}\textbf{0.740} \\
 &  & 60 & 0.910 & 0.909 & 0.012 & 0.215 & 0.056 & 0.178 & 0.862 & 0.302 & 0.710 & 0.450 \\
 &  & 50 & 0.641 & 0.641 & 0.008 & 0.209 & 0.052 & 0.171 & 0.860 & 0.293 & 0.680 & 0.280 \\ \cmidrule(l){2-13}
 & \multirow{3}{*}{SelectiveContext} & 80 & 0.919 & 0.918 & 0.016 & 0.218 & 0.066 & 0.186 & 0.863 & 0.327 & 0.740 & 0.540 \\
 &  & 60 & 0.901 & 0.899 & 0.007 & 0.197 & 0.046 & 0.155 & 0.859 & 0.271 & 0.720 & 0.270 \\
 &  & 50 & 0.638 & 0.638 & 0.006 & 0.196 & 0.043 & 0.153 & 0.857 & 0.264 & 0.720 & 0.150 \\ \midrule
 
\multirow{13}{*}{Llama-3 70B} & \multicolumn{1}{c}{N/A} & 100 & \cellcolor{mygray}0.953 & \cellcolor{mygray}0.953 & \cellcolor{mygray}0.020 & \cellcolor{mygray}0.235 & \cellcolor{mygray}0.073 & \cellcolor{mygray}0.196 & \cellcolor{mygray}0.876 & \cellcolor{mygray}0.333 & \cellcolor{mygray}0.874 & \cellcolor{mygray}0.919 \\ \cmidrule(l){2-13}
 &\multirow{3}{*}{GlobEnc} & 80 & 0.948 & 0.948 & \cellcolor{mycyan}\textbf{0.018} & \cellcolor{mycyan}\textbf{0.235} & \cellcolor{mycyan}\textbf{0.071} & \cellcolor{mycyan}\textbf{0.195} & \cellcolor{mycyan}\textbf{0.875} & \cellcolor{mycyan}\textbf{0.331} & \cellcolor{mycyan}\textbf{0.862} & 0.669 \\
 &  & 60 & 0.943 & 0.943 & 0.017 & 0.231 & 0.068 & 0.192 & 0.874 & 0.329 & 0.838 & 0.362 \\
 &  & 50 & 0.938 & 0.938 & 0.016 & 0.228 & 0.065 & 0.189 & 0.873 & 0.316 & 0.816 & 0.231 \\ \cmidrule(l){2-13}
 & \multirow{3}{*}{DecompX} & 80 & \cellcolor{mycyan}\textbf{0.949} & \cellcolor{mycyan}\textbf{0.949} & 0.017 & 0.226 & 0.065 & 0.187 & 0.870 & 0.316 & 0.839 & 0.818 \\
 &  & 60 & 0.891 & 0.891 & 0.015 & 0.215 & 0.057 & 0.176 & 0.868 & 0.292 & 0.797 & 0.587 \\
 &  & 50 & 0.839 & 0.837 & 0.014 & 0.209 & 0.054 & 0.171 & 0.866 & 0.281 & 0.770 & 0.409 \\ \cmidrule(l){2-13}
 & \multirow{3}{*}{LLMLingua2} & 80 & 0.938 & 0.937 & 0.015 & 0.231 & 0.069 & \cellcolor{mycyan}\textbf{0.195} & 0.867 & 0.329 & 0.860 & \cellcolor{mycyan}\textbf{0.860} \\
 &  & 60 & 0.889 & 0.890 & 0.017 & 0.229 & 0.067 & 0.191 & 0.866 & 0.293 & 0.835 & 0.590 \\
 &  & 50 & 0.644 & 0.643 & 0.012 & 0.225 & 0.057 & 0.185 & 0.865 & 0.265 & 0.797 & 0.440 \\ \cmidrule(l){2-13}
 & \multirow{3}{*}{SelectiveContext} & 80 & 0.920 & 0.919 & 0.015 & 0.230 & 0.068 & 0.193 & 0.868 & 0.329 & 0.855 & 0.810 \\
 &  & 60 & 0.887 & 0.887 & 0.009 & 0.214 & 0.058 & 0.173 & 0.867 & 0.292 & 0.815 & 0.440 \\
 &  & 50 & 0.634 & 0.634 & 0.010 & 0.203 & 0.045 & 0.163 & 0.866 & 0.268 & 0.740 & 0.220 \\ \bottomrule
\end{tabular}
\caption{Impact of prompt-compression on Llama-3 base models (8B and 70B) across tasks. Base and best prompt-reduced performance for each model are highlighted in Gray and {\textbf{bold}} Light Blue cells, respectively.}
\label{tab:resultsTableLlama}
\vspace{-4mm}
\end{table*}

\begin{table*}[h!]
\setlength{\tabcolsep}{3pt}
\renewcommand{\arraystretch}{0.9} 
\centering
\small
\begin{tabular}{@{}lcccccccccccc@{}}
\toprule
\multicolumn{1}{c}{\multirow{3}{*}{\textbf{Model}}} & \multirow{4}{*}{\textbf{\begin{tabular}[c]{@{}c@{}}Prompt\\ Reduction\\ Method\end{tabular}}} & \multirow{3}{*}{\textbf{$k\%$}} & \multicolumn{10}{c}{\textbf{Tasks}} \\ \cmidrule(l){4-13}
\multicolumn{1}{c}{} &  &  & \multicolumn{2}{c}{\textbf{CLS}} & \multicolumn{6}{c}{\textbf{SUM}} & \textbf{QA} & \textbf{RSN} \\ \cmidrule(lr){4-5} \cmidrule(lr){6-11} \cmidrule(lr){12-12} \cmidrule(lr){13-13}
\multicolumn{1}{c}{} &  &  & \textbf{Acc} & \textbf{F1} & \textbf{BLEU} & \textbf{R-1} & \textbf{R-2} & \textbf{R-L} & \textbf{BERT} & \textbf{METEOR} & \textbf{Acc} & \textbf{pass@1} \\ \midrule
 
\multirow{13}{*}{GPT-3.5} & \multicolumn{1}{c}{N/A} & \multicolumn{1}{c}{100} & \cellcolor{mygray}0.949 & \cellcolor{mygray}0.949 & \cellcolor{mygray}0.039 & \cellcolor{mygray}0.282 & \cellcolor{mygray}0.093 & \cellcolor{mygray}0.237 & \cellcolor{mygray}0.889 & \cellcolor{mygray}0.359 & \cellcolor{mygray}0.779 & \cellcolor{mygray}0.772 \\ \cmidrule(l){2-13}
 &\multirow{3}{*}{GlobEnc} & 80 & \cellcolor{mycyan}\textbf{0.945} & \cellcolor{mycyan}\textbf{0.945} & 0.017 & 0.176 & 0.056 & 0.146 & 0.874 & 0.291 & \cellcolor{mycyan}\textbf{0.753} & 0.498 \\
 &  & 60 & 0.925 & 0.925 & 0.015 & 0.172 & 0.052 & 0.141 & 0.874 & 0.285 & 0.716 & 0.264 \\
 &  & 50 & 0.918 & 0.918 & 0.014 & 0.170 & 0.049 & 0.140 & 0.873 & 0.278 & 0.705 & 0.158 \\ \cmidrule(l){2-13}
 & \multirow{3}{*}{DecompX} & 80 & 0.942 & 0.942 & \cellcolor{mycyan}\textbf{0.036} & 0.268 & 0.084 & 0.225 & \cellcolor{mycyan}\textbf{0.888} & 0.337 & 0.704 & 0.660 \\
 &  & 60 & 0.724 & 0.704 & 0.031 & 0.253 & 0.073 & 0.210 & 0.885 & 0.311 & 0.648 & 0.419 \\
 &  & 50 & 0.642 & 0.595 & 0.027 & 0.241 & 0.065 & 0.200 & 0.882 & 0.291 & 0.619 & 0.288 \\ \cmidrule(l){2-13}
 & \multirow{3}{*}{LLMLingua2} & 80 & 0.933 & 0.932 & 0.035 & 0.265 & 0.092 & 0.221 & 0.886 & 0.325 & 0.750 & \cellcolor{mycyan}\textbf{0.700} \\
 &  & 60 & 0.910 & 0.909 & 0.032 & 0.265 & 0.084 & 0.220 & 0.883 & 0.312 & 0.740 & 0.460 \\
 &  & 50 & 0.710 & 0.708 & 0.029 & 0.261 & 0.079 & 0.218 & 0.880 & 0.323 & 0.700 & 0.300 \\ \cmidrule(l){2-13}
 & \multirow{3}{*}{SelectiveContext} & 80 & 0.920 & 0.919 & 0.031 & \cellcolor{mycyan}\textbf{0.277} & \cellcolor{mycyan}\textbf{0.099} & \cellcolor{mycyan}\textbf{0.234} & 0.884 & \cellcolor{mycyan}\textbf{0.366} & 0.751 & 0.650 \\
 &  & 60 & 0.910 & 0.908 & 0.025 & 0.244 & 0.072 & 0.199 & 0.881 & 0.302 & 0.720 & 0.360 \\
 &  & 50 & 0.681 & 0.680 & 0.022 & 0.228 & 0.063 & 0.180 & 0.879 & 0.283 & 0.650 & 0.200 \\ \midrule
 
\multirow{13}{*}{\begin{tabular}[c]{@{}l@{}}Gemini 2.0\\ Flash\\ Thinking\end{tabular}} & \multicolumn{1}{c}{N/A} & \multicolumn{1}{c}{100} & \cellcolor{mygray}0.952 & \cellcolor{mygray}0.952 & \cellcolor{mygray}0.034 & \cellcolor{mygray}0.262 & \cellcolor{mygray}0.081 & \cellcolor{mygray}0.219 & \cellcolor{mygray}0.885 & \cellcolor{mygray}0.345 & \cellcolor{mygray}0.880 & \cellcolor{mygray}0.956 \\ \cmidrule(l){2-13}
 & \multirow{3}{*}{GlobEnc} & 80 & 0.947 & 0.947 & 0.031 & 0.252 & 0.081 & 0.212 & 0.882 & 0.344 & \cellcolor{mycyan}\textbf{0.879} & 0.704 \\
 &  & 60 & 0.934 & 0.934 & 0.029 & 0.247 & 0.077 & 0.208 & 0.881 & 0.335 & 0.846 & 0.423 \\
 &  & 50 & 0.920 & 0.919 & 0.026 & 0.239 & 0.071 & 0.199 & 0.879 & 0.322 & 0.827 & 0.277 \\ \cmidrule(l){2-13}
 & \multirow{3}{*}{DecompX} & 80 & 0.855 & 0.852 & 0.031 & 0.251 & 0.075 & 0.209 & \cellcolor{mycyan}\textbf{0.883} & 0.328 & 0.856 & 0.856 \\
 &  & 60 & 0.713 & 0.690 & 0.028 & 0.236 & 0.068 & 0.194 & 0.878 & 0.307 & 0.795 & 0.665 \\
 &  & 50 & 0.627 & 0.571 & 0.024 & 0.225 & 0.059 & 0.185 & 0.876 & 0.283 & 0.774 & 0.463 \\ \cmidrule(l){2-13}
 & \multirow{3}{*}{LLMLingua2} & 80 & 0.940 & 0.939 & 0.030 & 0.248 & 0.079 & \cellcolor{mycyan}\textbf{0.242} & 0.882 & \cellcolor{mycyan}\textbf{0.390} & 0.790 & \cellcolor{mycyan}\textbf{0.910} \\
 &  & 60 & 0.920 & 0.919 & 0.029 & 0.235 & 0.076 & 0.230 & 0.877 & 0.367 & 0.840 & 0.640 \\
 &  & 50 & 0.901 & 0.899 & 0.023 & 0.224 & 0.072 & 0.214 & 0.873 & 0.326 & 0.810 & 0.460 \\ \cmidrule(l){2-13}
 & \multirow{3}{*}{SelectiveContext} & 80 & \cellcolor{mycyan}\textbf{0.950} & \cellcolor{mycyan}\textbf{0.949} & \cellcolor{mycyan}\textbf{0.043} & \cellcolor{mycyan}\textbf{0.270} & \cellcolor{mycyan}\textbf{0.096} & 0.225 & 0.880 & 0.366 & 0.790 & 0.800 \\
 &  & 60 & 0.902 & 0.901 & 0.032 & 0.247 & 0.079 & 0.206 & 0.875 & 0.318 & 0.800 & 0.400 \\
 &  & 50 & 0.897 & 0.897 & 0.025 & 0.235 & 0.064 & 0.192 & 0.872 & 0.299 & 0.750 & 0.200 \\ \midrule
 
\multirow{13}{*}{o3-mini} & \multicolumn{1}{c}{N/A} & \multicolumn{1}{c}{100} & \cellcolor{mygray}0.957 & \cellcolor{mygray}0.957 & \cellcolor{mygray}0.023 & \cellcolor{mygray}0.221 & \cellcolor{mygray}0.065 & \cellcolor{mygray}0.182 & \cellcolor{mygray}0.860 & \cellcolor{mygray}0.297 & \cellcolor{mygray}0.845 & \cellcolor{mygray}0.961 \\ \cmidrule(l){2-13}
 & \multirow{3}{*}{GlobEnc} & 80 & 0.956 & 0.956 & \cellcolor{mycyan}\textbf{0.020} & 0.216 & 0.060 & 0.176 & \cellcolor{mycyan}\textbf{0.859} & 0.290 & \cellcolor{mycyan}\textbf{0.826} & 0.724 \\
 &  & 60 & 0.941 & 0.941 & 0.019 & 0.212 & 0.059 & 0.173 & 0.858 & 0.282 & 0.802 & 0.462 \\
 &  & 50 & 0.932 & 0.932 & 0.018 & 0.204 & 0.055 & 0.166 & 0.857 & 0.272 & 0.785 & 0.332 \\ \cmidrule(l){2-13}
 & \multirow{3}{*}{DecompX} & 80 & 0.842 & 0.839 & \cellcolor{mycyan}\textbf{0.020} & 0.208 & 0.056 & 0.170 & 0.858 & 0.273 & 0.787 & 0.850 \\
 &  & 60 & 0.727 & 0.707 & 0.018 & 0.195 & 0.049 & 0.159 & 0.854 & 0.253 & 0.724 & 0.679 \\
 &  & 50 & 0.641 & 0.593 & 0.017 & 0.187 & 0.045 & 0.152 & 0.853 & 0.236 & 0.686 & 0.533 \\ \cmidrule(l){2-13}
 & \multirow{3}{*}{LLMLingua2} & 80 & \cellcolor{mycyan}\textbf{0.970} & \cellcolor{mycyan}\textbf{0.969} & 0.019 & 0.213 & 0.057 & 0.168 & 0.858 & 0.274 & 0.820 & \cellcolor{mycyan}\textbf{0.920} \\
 &  & 60 & 0.930 & 0.928 & 0.017 & 0.204 & 0.052 & 0.158 & 0.856 & 0.270 & 0.820 & 0.700 \\
 &  & 50 & 0.910 & 0.908 & 0.016 & 0.203 & 0.047 & 0.155 & 0.853 & 0.262 & 0.790 & 0.530 \\ \cmidrule(l){2-13}
 & \multirow{3}{*}{SelectiveContext} & 80 & 0.930 & 0.929 & 0.019 & \cellcolor{mycyan}\textbf{0.232} & \cellcolor{mycyan}\textbf{0.067} & \cellcolor{mycyan}\textbf{0.186} & 0.856 & \cellcolor{mycyan}\textbf{0.312} & 0.810 & 0.830 \\
 &  & 60 & 0.910 & 0.908 & 0.017 & 0.203 & 0.052 & 0.157 & 0.854 & 0.269 & 0.770 & 0.460 \\
 &  & 50 & 0.671 & 0.670 & 0.016 & 0.186 & 0.043 & 0.145 & 0.852 & 0.238 & 0.710 & 0.280 \\ \bottomrule
\end{tabular}
\caption{Impact of prompt-compression on proprietary models (GPT-3.5, Gemini 2.0 Flash Thinking, and o3-mini) across tasks. Base and best prompt-reduced performance for each model are highlighted in Gray and {\textbf{bold}} Light Blue cells, respectively.}
\label{tab:resultsTableProprietary}
\vspace{-4mm}
\end{table*}

The proof of Theorem \ref{thm:del-bound} is provided in Appendix \S\ref{sec:proof}. Theorem \ref{thm:del-bound} establishes the stability of token deletion by upper-bounding the deletion effect.
\paragraph{Retention Relationship.}
Following \S\ref{sec:topkFiltering}, we define the deleted set $D_k := \{\pi(p+1),\dots,\pi(n)\}$ and the frugalized sequence $F_k := T\setminus D_k$.
\begin{corollary}[Performance \textit{vs.}\ $k\%$ retention]
\label{cor:del-k}
Under {Assumptions~\ref{assump:del} and~\ref{assump:pairwiseBound}}, with $q=n-p$,
{\setlength{\abovedisplayskip}{2pt}
\setlength{\belowdisplayskip}{2pt}
{\small
\begin{align}
\nonumber\ell_\tau(F_k,y;\theta)-\ell_\tau&(T,y;\theta)
\;\le\;
C_\tau \sum_{j=p+1}^{n} s_{\pi(j)}
\;\\
&+\;\frac{\gamma_\tau}{2}\,(n-p)(n-p-1).
\label{eq:k-del}
\end{align}
}
}
\end{corollary}  
{Corollary~\ref{cor:del-k} follows directly by substituting $D$ in Theorem~\ref{thm:del-bound} and expanding $f(D_k)$ via Eq.~\ref{eq:del-set-fn}. The bound also generalizes to the dataset level. Given that $(T,y)\sim\mathcal{D}_\tau$ denotes the test distribution for task $\tau$, we define the risk as the expected loss,}
{\setlength{\abovedisplayskip}{2pt}
\setlength{\belowdisplayskip}{2pt}
\begin{align}
    {R_\tau(\theta) := \mathbb{E}_{(T,y)\sim\mathcal{D}_\tau}[\ell_\tau(T,y;\theta)],}
\end{align}
}
{and taking expectations on both sides of Eq.~\ref{eq:k-del} bounds the increase in risk under frugalization by the expected value of its right-hand side.}
\paragraph{Interpretation.} Because we delete the lowest-saliency tokens, $\sum_{j=p+1}^{n} s_{\pi(j)}$ is \emph{minimized} over all subsets of size $n-p$. Top-$k$ filtering is the optimal one-shot deletion strategy with respect to the bound in Eq.~\ref{eq:k-del}. The bound also separates two failure modes. The linear term $C_\tau \sum_{j=p+1}^{n} s_{\pi(j)}$ captures \emph{information loss}{, which is controllable by the ranking and small when} the saliency distribution is concentrated on a few tokens. The quadratic term $\tfrac{\gamma_\tau}{2}(n-p)(n-p-1)$ captures \emph{compositional breakdown}, which is uncontrollable by the ranking and dominant when $n-p$ is large. This decomposition predicts the experimental pattern observed in \S\ref{sec:resultAnalysis}. Tasks with sparse saliency (classification, question answering) tolerate aggressive deletion, while tasks with dense token dependencies (mathematical reasoning) are interaction-limited and degrade quadratically as $k$ shrinks.

\begin{figure*}[ht]
    \centering
    \includegraphics[width=\linewidth]{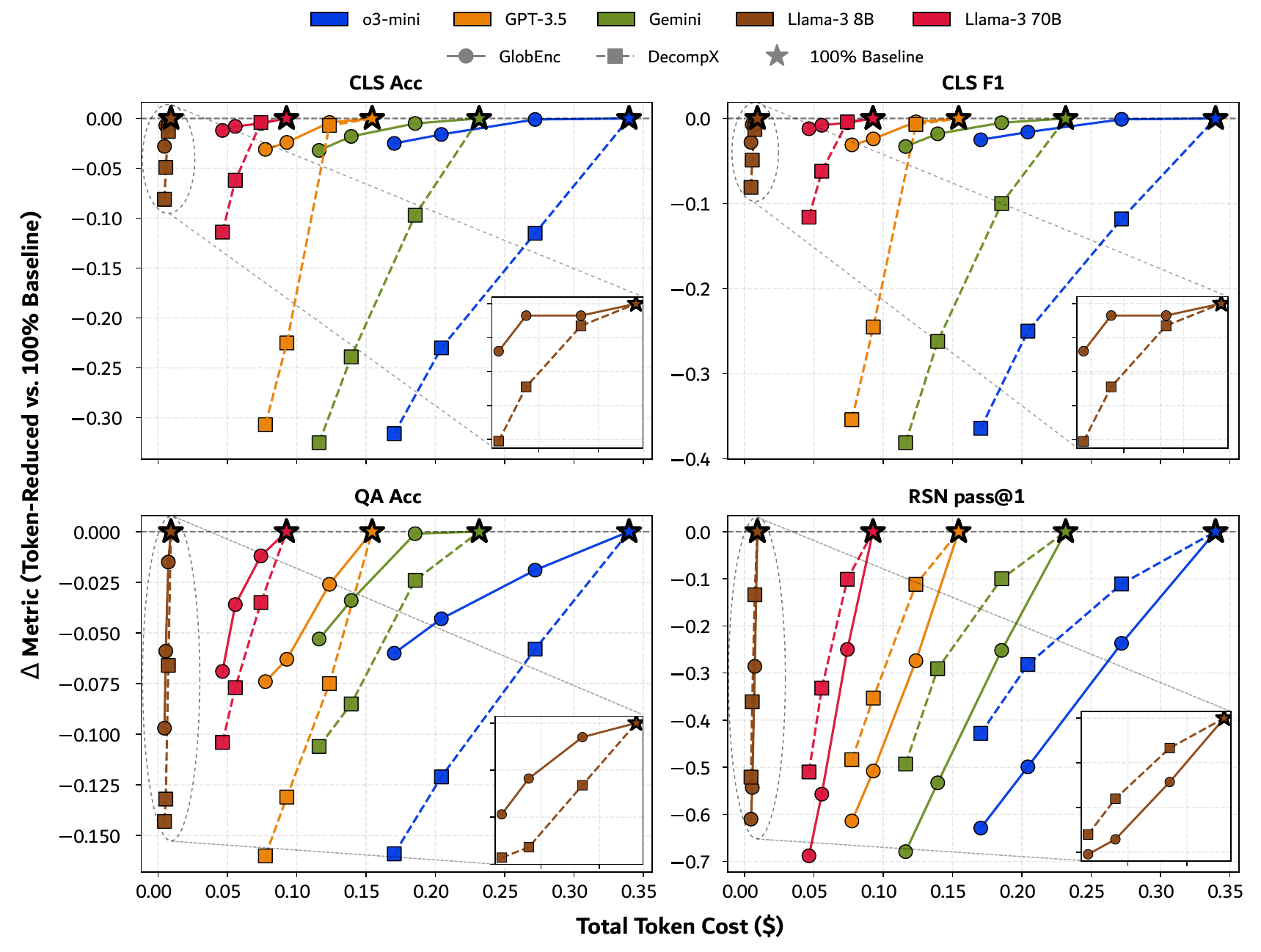} 
    \caption{{Performance difference between the token-reduced prompts and the $100\%$-token baseline, plotted against the total token cost, for the metrics of three tasks: text classification (CLS), question answering (QA), and reasoning (RSN). Insets zoom into the low-cost Llama-3 8B region.}}
    \label{fig:scatterClassificationMetrics}
\end{figure*}

\section{{Results and Analysis}} 
\label{sec:resultAnalysis}
{Having established the performance bounds and their relationship with the retention $k\%$ in \S\ref{sec:theoreticalBounds}, we now provide empirical evidence in this section, with additional experimental details reported in Appendix \S\ref{sec:app_experimentalDetails}.}

\subsection{{Experimental Setup}}
\label{subsec:experimentalSetup}
{We set the temperature to $0$ for all inference runs, \textit{i.e.}, greedy decoding, ensuring deterministic and reproducible generations. For the frugalized setup, we experiment by retaining the top $k\in\{50,60,80\}\%$ of the input tokens. We also include the baseline prompt, using $100\%$ of the tokens to set a proper frame of reference for the performance delta across all models. This setup enables us to investigate whether low-attribution filler tokens can be excluded with minimal degradation in task performance, thereby reducing API costs and carbon footprint.} Additional experimental details on the tasks (\S\ref{subsec:tasks}), datasets (\S\ref{subsec:dataset}), models (\S\ref{subsec:models}), and evaluation metrics (\S\ref{subsec:evaluationMetrics}) have been reported in the Appendix \S\ref{sec:app_experimentalDetails}.

\subsection{Performance across Tasks}
\begin{figure*}[ht]
    \centering
    \includegraphics[width=\linewidth]{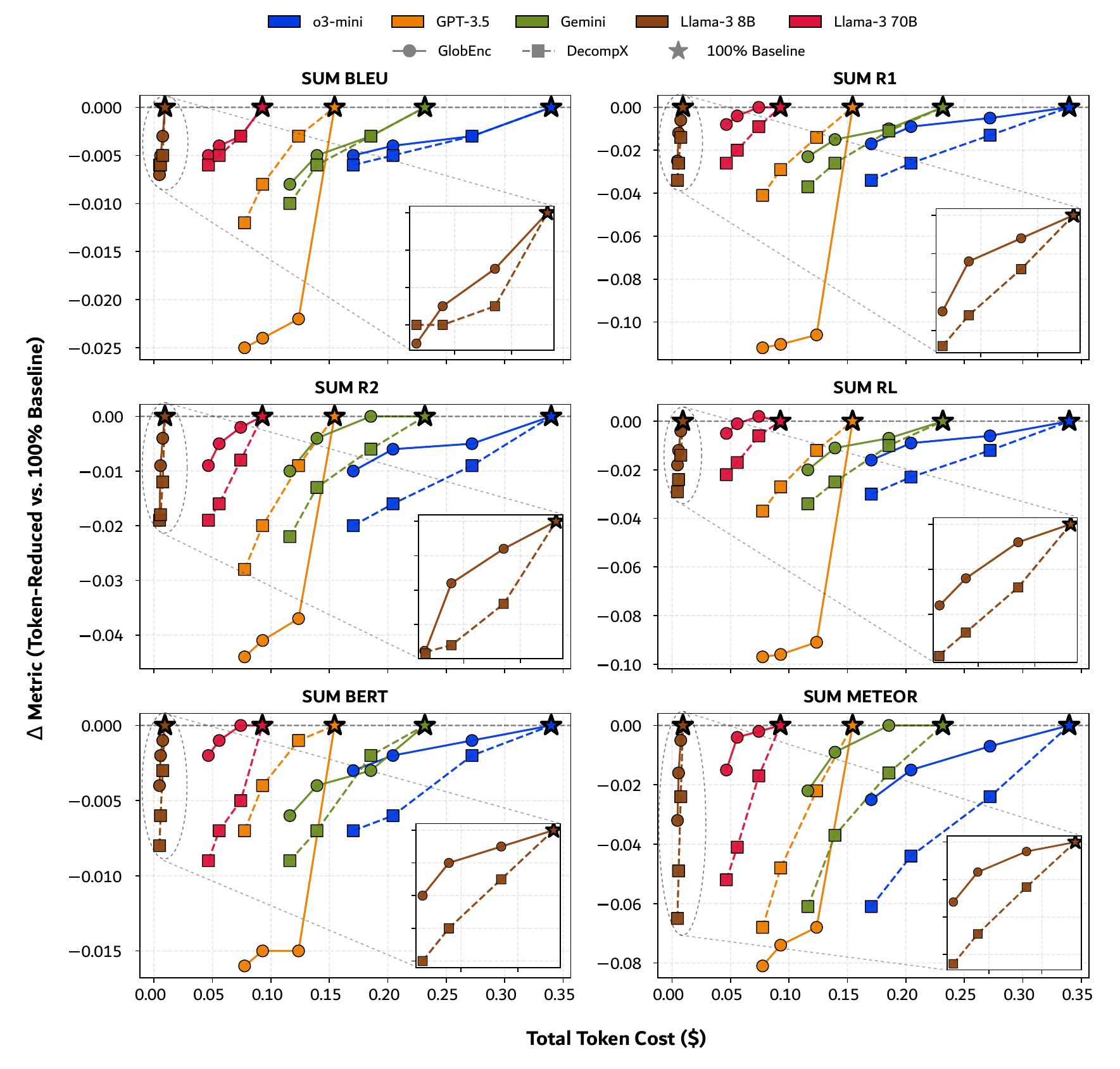}
    \caption{{Performance difference between the token-reduced prompts and the $100\%$-token baseline, plotted against the total token cost, across the six summarization metrics. Insets zoom into the low-cost Llama-3 8B region.}}
    \label{fig:scatterSummarizationMetrics}
\end{figure*}

Tables {\ref{tab:resultsTableLlama} and \ref{tab:resultsTableProprietary}} reveal that performance degradation becomes more pronounced as the retention threshold decreases, particularly below $60\%$ token retention, with distinct patterns emerging across tasks. {For instance, at $50\%$ retention, sentiment classification accuracy drops by {$1.5$--$3.2$} points with GlobEnc but by $8$--$33$ points with DecompX, commonsense QA accuracy drops by $5$--$16$ points, and summarization metrics such as ROUGE-1 decline by $3$--$15\%$ relative to the baseline (with GPT-3.5 under GlobEnc as an outlier at roughly $40\%$), as seen in Figures~\ref{fig:scatterClassificationMetrics} and~\ref{fig:scatterSummarizationMetrics}.} The performance decline is noticeably smaller for higher values of $k$, \textit{e.g.,} GlobEnc using $80\%$ tokens achieves close to the baseline performance in classification and QA tasks. 
 
Mathematical reasoning shows the steepest fall, {with pass@1 scores dropping by $43$--$69$ points at $50\%$ retention (falling to as low as $14.6\%$ for Llama-3 8B)}, reflecting the loss of critical numerical and logical connectors. We hypothesize that this degradation is more pronounced for reasoning tasks because they rely on the sequential integrity of tokens and numerical data. Low $k\%$ thresholds exacerbate this by discarding essential context, amplifying errors in tasks requiring unbroken symbolic chains, while simpler classification and summarization tasks generally tolerate greater omission by inferring from salient cues.

\definecolor{mygray}{gray}{0.9}       
\definecolor{mycyan}{RGB}{189, 224, 252} 

\begin{table*}[t]
\centering
\resizebox{0.98\textwidth}{!}{
\begin{tabular}{@{}ccccccccccccc@{}}
\toprule
\multirow{5}{*}{\textbf{\begin{tabular}[c]{@{}c@{}}Attribution\\Type\end{tabular}}}  & \multirow{5}{*}{\textbf{\begin{tabular}[c]{@{}c@{}}Token\\ Attribution\\Method\end{tabular}}} & \multirow{4}{*}{\textbf{$k\%$}} & \multicolumn{10}{c}{\textbf{Tasks}} \\ \cmidrule(l){4-13} 
\multicolumn{1}{c}{} &  &  & \multicolumn{2}{c}{\textbf{CLS}} & \multicolumn{6}{c}{\textbf{SUM}} & \textbf{QA} & \textbf{RSN} \\ \cmidrule(lr){4-5} \cmidrule(lr){6-11} \cmidrule(lr){12-12} \cmidrule(lr){13-13}
\multicolumn{1}{c}{} &  &  & \multicolumn{2}{c}{\textbf{\texttt{o3-mini}}} & \multicolumn{6}{c}{\textbf{\texttt{GPT-3.5}}} & {\textbf{\texttt{Gemini-2.0 FT}}} & \textbf{\texttt{o3-mini}} \\
\cmidrule(lr){4-5} \cmidrule(lr){6-11} \cmidrule(lr){12-12} \cmidrule(lr){13-13}
\multicolumn{1}{c}{} &  &  & \textbf{Acc} & \textbf{F1} & \textbf{BLEU} & \textbf{R-1} & \textbf{R-2} & \textbf{R-L} & \textbf{BERT} & \textbf{METEOR} & \textbf{Acc} & \textbf{pass@1} \\ \midrule

 \multicolumn{2}{c}{N/A}& 100 & \cellcolor{mygray}0.955 & \cellcolor{mygray}0.955 & \cellcolor{mygray}0.039 & \cellcolor{mygray}0.282 & \cellcolor{mygray}0.093 & \cellcolor{mygray}0.237 & \cellcolor{mygray}0.889 & \cellcolor{mygray}0.359 & \cellcolor{mygray}0.880 & \cellcolor{mygray}0.961 \\ \cmidrule(l){1-13}
\multirow{7}{*}{Random} 
 & \multirow{3}{*}{GlobEnc} & 80 & \cellcolor{mycyan}\textbf{0.951} & \cellcolor{mycyan}\textbf{0.951} & 0.030 & \cellcolor{mycyan}\textbf{0.261} & 0.078 & 0.217 & 0.886 & \cellcolor{mycyan}\textbf{0.329} & \cellcolor{mycyan}\textbf{0.850} & 0.373 \\
 
 &  & 60 & 0.893 & 0.893 & 0.022 & 0.240 & 0.064 & 0.197 & 0.883 & 0.293 & 0.761 & 0.103 \\
 
 &  & 50 & 0.886 & 0.886 & 0.020 & 0.223 & 0.053 & 0.182 & 0.880 & 0.267 & 0.713 & 0.058 \\ \cmidrule(l){2-13}
 
 & \multirow{3}{*}{DecompX} & 80 & 0.942 & 0.942 & \cellcolor{mycyan}\textbf{0.033} & 0.259 & \cellcolor{mycyan}\textbf{0.080} & \cellcolor{mycyan}\textbf{0.218} & \cellcolor{mycyan}\textbf{0.887} & 0.328 & 0.837 & \cellcolor{mycyan}\textbf{0.381} \\

 &  & 60 & 0.901 & 0.901 & 0.024 & 0.237 & 0.064 & 0.196 & 0.882 & 0.289 & 0.779 & 0.114 \\
 
 &  & 50 & 0.893 & 0.893 & 0.022 & 0.223 & 0.055 & 0.182 & 0.879 & 0.262 & 0.734 & 0.011 \\ \cmidrule(l){1-13}

\multirow{7}{*}{Bottom}
 & \multirow{3}{*}{GlobEnc} & 80 & \cellcolor{mycyan}\textbf{0.914} & \cellcolor{mycyan}\textbf{0.914} & 0.026 & 0.234 & 0.067 & 0.195 & 0.883 & 0.289 & 0.764 & 0.040 \\
 
 &  & 60 & 0.760 & 0.759 & 0.012 & 0.139 & 0.030 & 0.117 & 0.865 & 0.158 & 0.672 & 0.028 \\
 
 &  & 50 & 0.626 & 0.616 & 0.007 & 0.084 & 0.012 & 0.072 & 0.850 & 0.091 & 0.625 & 0.014 \\ \cmidrule(l){2-13}

 & \multirow{3}{*}{DecompX} & 80 & 0.857 & 0.854 & \cellcolor{mycyan}\textbf{0.031} & \cellcolor{mycyan}\textbf{0.247} & \cellcolor{mycyan}\textbf{0.073} & \cellcolor{mycyan}\textbf{0.205} & \cellcolor{mycyan}\textbf{0.884} & \cellcolor{mycyan}\textbf{0.302} & \cellcolor{mycyan}\textbf{0.809} & \cellcolor{mycyan}\textbf{0.045} \\

 &  & 60 & 0.755 & 0.740 & 0.022 & 0.208 & 0.053 & 0.170 & 0.876 & 0.249 & 0.714 & 0.021 \\
 
 &  & 50 & 0.696 & 0.666 & 0.020 & 0.192 & 0.047 & 0.156 & 0.873 & 0.225 & 0.670 & 0.014 \\

\bottomrule
\end{tabular}%
}
\caption{Impact of retaining random and bottom $k\%$ tokens using the two variants of \textsc{FrugalPrompt} with the best performing models for each task: text classification (CLS), text summarization (SUM), question answering (QA), and {mathematical} reasoning (RSN). {Base and best performance for each attribution type are highlighted in Gray and \textbf{bold} Light Blue cells, respectively.}}
\label{tab:ablationResults}
\end{table*}

\subsection{Choice of Attribution Method}
Following {Tables~\ref{tab:resultsTableLlama} and~\ref{tab:resultsTableProprietary} and Figure~\ref{fig:scatterClassificationMetrics}}, for classification and QA tasks, GlobEnc outperforms DecompX in nearly all settings, with the only exception of reasoning. DecompX's subword-level attribution, which includes linear activations, may help retain more important tokens for reasoning but might overwhelm models on other tasks. For summarization (Fig. \ref{fig:scatterSummarizationMetrics}), GlobEnc usually outperforms DecompX, except for the GPT-3.5 model. While DecompX is reported to produce \textit{better} attribution scores \cite{modarressi-etal-2023-decompx}, most models find the token sequence filtered by {attention-based aggregation}, such as GlobEnc, to be more useful for retaining performance across most tasks.
\subsection{{Comparison with Baselines}}
{Tables~\ref{tab:resultsTableLlama} and~\ref{tab:resultsTableProprietary} also report two prompt compression baselines, LLMLingua-2 \cite{pan-etal-2024-llmlingua} and Selective Context \cite{Li2023CompressingCT}, at matched retention ratios. For classification and QA, GlobEnc is the most robust method under aggressive compression: at $50\%$ retention, the sentiment accuracy of both baselines falls to $0.63$--$0.71$ on the Llama-3 models and GPT-3.5, whereas GlobEnc retains $0.92$--$0.94$ across all five models, and GlobEnc yields the best QA accuracy in 11 of the 15 model--retention settings. Conversely, LLMLingua-2 is the strongest method for mathematical reasoning at $80\%$ retention on all five models, and Selective Context attains the best summarization ROUGE-1 at $80\%$ retention on the proprietary models, while GlobEnc remains the best summarizer on the Llama-3 models. Overall, training-free saliency filtering is most advantageous for tasks whose saliency mass is concentrated in a few content tokens, whereas sequentially dependent tasks favor compressors that preserve token continuity.}
\begin{table}[t]
\centering
\small
\setlength{\tabcolsep}{5pt}
\begin{tabular}{lcc}
\toprule
\textbf{Model} & \textbf{Input (\$)} & \textbf{Output (\$)} \\
\midrule
\texttt{Llama-3 8B} & 0.03 & 0.06 \\
\texttt{Llama-3 70B} & 0.30 & 0.40 \\
\texttt{GPT-3.5} & 0.50 & 1.50 \\
\texttt{Gemini-2.0 FT} & 0.10 & 0.40 \\
\texttt{o3-mini} & 1.10 & 4.40 \\
\bottomrule
\end{tabular}
\caption{{Cost per 1 million input and output tokens for each model. The Gemini 2.0 Flash Thinking (FT) model has output cost fluctuations due to the number of thinking tokens produced.}}
\label{tab:model_costs}
\end{table}
\subsection{Performance \textit{vs.} Cost}
The cost per million tokens for each model was taken from the respective provider's website, at the time of writing, and is reported in Table \ref{tab:model_costs}. Note that the price is subject to change. The cost of the open-source Llama models was estimated using the lowest available rate on OpenRouter\footref{fn:openrouter}.

Figures \ref{fig:scatterClassificationMetrics} and \ref{fig:scatterSummarizationMetrics} show the performance difference between the reduced set of tokens and $100\%$ tokens for all the models. For sentiment analysis and question answering, we observe that using a reduced set of tokens generally maintains performance and shows a gentler decline as we decrease the cost. However, reducing tokens sharply degrades performance on reasoning tasks, making token reduction unsuitable. The results are mixed for summarization tasks. While most models have a softer performance decline across the metrics, {models} such as GPT-3.5 using GlobEnc experience rapid performance degradation. 

% ...

\subsection{Performance across Models}
Figures \ref{fig:scatterClassificationMetrics} and \ref{fig:scatterSummarizationMetrics} show that for smaller models, token attribution offers negligible cost savings relative to the sharper performance degradation it entails. Consequently, token reduction is more practical for larger and costlier models. Legacy models such as GPT-3.5 often display abrupt drops in performance, \textit{e.g.}, in summarization using GlobEnc, necessitating greater caution when reducing tokens.


\subsection{{Random \& Bottom $k$ tokens}} 
We select random $k\%$ and the bottom $k\%$ tokens (based on attribution scores), while preserving the text order, to establish a baseline for our token reduction method. 
As shown in \autoref{tab:ablationResults}, classification, summarization, and question answering exhibit {surprisingly strong retention in performance, particularly at $80\%$ retention, where even the bottom-$k\%$ tokens retain $91.4\%$ classification accuracy and $76$--$81\%$ QA accuracy}. This strongly suggests potential task contamination \cite{li2024task}, since the ground-truth answers for the test sets are readily available online. Reasoning tasks experience a {far sharper decline (pass@1 below $0.4$ even with random $80\%$ tokens and below $0.05$ with bottom-$k\%$ tokens)}, indicating a genuine reliance on contextual information to solve the task.






\section{Conclusion}
The surging cost of LLMs has often made {them} inaccessible to people in under-resourced environments, while increasing carbon footprint and {inference-time} latency. Our work aims to bridge this gap by introducing prompt compression at the token-attribution level, thereby reducing the cost of LLM inference and making it accessible to a wider range of users. We also identify potential instances of task contamination to consider when evaluating models. We wish to explore adaptive frugalization strategies that dynamically adjust token retention based on task complexity and model confidence. Our results suggest that retention is better treated as a task-conditioned judgment than a global setting. We wish to adaptively tune token retention using task complexity and model confidence, targeting long-context and agentic regimes where verbosity exacts the highest toll. Ultimately, the question of how much language a model requires is inseparable from the question of how much meaning it can supply unaided: every token we elide is a wager on what the listener already knows, and in learning to compress our prompts, we are learning to articulate what we believe our machines can understand without being told.
\section*{Limitations}

\paragraph{Attribution as a proxy for causal necessity.}
Our pipeline presumes that the salience an encoder assigns to a token is a faithful surrogate for the harm incurred by deleting that token from the prompt of a far larger autoregressive model. Assumption~\ref{assump:del} states this premise explicitly, but it relates two distinct systems: the scores are produced by a 110M-parameter BERT-class encoder, whereas the deletion effects of interest are realized in Llama-3, GPT-3.5, Gemini 2.0 Flash Thinking, and o3-mini. We neither verify this cross-model correspondence empirically nor estimate the task constants $C_\tau$ and $\gamma_\tau$. Theorem~\ref{thm:del-bound} and Corollary~\ref{cor:del-k} should accordingly be read as a \emph{conditional} guarantee, \textit{i.e.}, an account of why top-$k$ selection is the optimal one-shot deletion rule \emph{when} faithfulness and bounded interaction hold, rather than as a certificate of stability for the models we evaluate. Wherever salience and causal necessity diverge, the ranking will discard tokens the target model genuinely requires, and the bound will remain too loose to predict the retention threshold at which compression ceases to be safe.

\paragraph{Reliance on task-aligned encoders.}
The scoring function $\varphi_\tau$ is task-specific by construction. \shortName{} is therefore training-free with respect to the \emph{inference} model, in that no gradient update is applied to the LLM whose context we compress, but it is not assumption-free with respect to the \emph{compressor}, since it presupposes an encoder whose objective is aligned with the downstream task. To the extent that such alignment is procured through task-specific supervision, our comparison against compressors that operate without that supervision should be interpreted with this asymmetry in view. The requirement also narrows applicability in open-ended, multi-task, and instruction-following regimes, where no single task label exists on which to condition attribution.

\paragraph{An impoverished account of redundancy.}
Our linguistic motivation treats function words and discourse expressions as carriers of negligible semantic weight, yet this conflates propositional content with syntactic and pragmatic function. Determiners encode definiteness, auxiliaries encode tense, aspect, and modality, negation particles invert truth conditions, and discourse markers such as \textit{``by the way''} regulate information structure and speaker stance while contributing little propositionally. Retention in our framework is governed by a single scalar per token, an ordering that admits no principled distinction among these levels of linguistic organization, and we do not measure separately what is lost when pragmatic scaffolding, rather than lexical content, is removed.

\paragraph{The contamination interpretation is suggestive, not conclusive.}
The performance retained under random-$k\%$ and bottom-$k\%$ selection (\autoref{tab:ablationResults}) is consistent with task contamination but does not isolate it. Ordinary robustness to elided context, label priors, answer-choice artifacts, residual lexical cues, and world knowledge acquired during pretraining are all plausible competing explanations, and our design cannot adjudicate among them. A decisive test would require contamination-controlled evidence, such as newly constructed or privately held test sets, or systematically perturbed variants of the same items, none of which we provide. The same ambiguity qualifies our principal findings in the opposite direction: insofar as these benchmarks are contaminated, the retention--performance curves partly reflect memorization rather than successful semantic preservation, which weakens any inference about how much meaning survives frugalization.

\paragraph{Scope of the empirical evaluation.}
Our benchmarks are standard but modest in length, and they are not the settings in which contextual overhead is most burdensome: protracted multi-turn dialogue, retrieval-augmented contexts, and agentic pipelines bearing extensive system prompts and tool specifications. The sentence-level chunking described in Appendix~\ref{app:longInputs} admits inputs of arbitrary length, but scores each sentence in isolation, leaving cross-sentential and document-level dependencies invisible to the attribution encoder. All experiments are conducted in English, on $1000$ samples per dataset, under deterministic decoding at a single temperature, so multilingual behavior and sensitivity to the decoding configuration remain uncharacterized.

\paragraph{Partial accounting of efficiency.}
Our cost analysis is denominated in provider input-token prices and thus captures only one component of end-to-end expenditure. It excludes the compute and wall-clock latency of the attribution pass, the cost of output tokens (appreciable for reasoning models that emit long deliberative traces), and the mismatch between the encoder's tokenizer and that of the target model, which entails that retained subwords do not correspond one-to-one to billed LLM tokens. Our appeals to inference latency and carbon footprint are therefore plausible extrapolations from token savings rather than directly measured quantities. As noted in \S\ref{sec:resultAnalysis}, the savings are already slender for smaller models relative to the degradation incurred, and a complete end-to-end audit may well reveal a net loss in that regime.

\paragraph{Safety and instruction integrity.}
Because the ranking is indifferent to illocutionary force, compression may excise negation, quantifiers, numerical constraints, or the instruction-defining and safety-critical clauses that determine what the model is being asked to do. Our reasoning results already demonstrate the fragility of sequential and symbolic dependencies under deletion; the analogous hazard for instruction-following and safety-conditioned prompts, including negation-sensitive inputs, multi-instruction prompts, and adversarial injections, is untested here. Pending such an evaluation, we advise against deploying \shortName{} on prompts whose correctness or safety rests upon the integrity of every constraint.

\section*{Ethical Considerations}
As identified in our work, performance retention in certain tasks might stem from the model's exposure to benchmark data during pretraining. This raises concerns about the validity of LLM evaluation and the need for contamination-free datasets. Prompt compression can remove safety-critical tokens and should therefore be used with caution.

\section*{Acknowledgements}
We convey our heartfelt gratitude to the anonymous
reviewers for their constructive criticisms and insightful feedback which were conducive to the improvement of the research work
outlined in this paper. We also appreciate the Systems and Software Lab (SSL) of the Islamic University of Technology (IUT) and The University of Queensland for the generous provision of computing resources during the course of
this project. Syed Rifat Raiyan, in particular, wants
to thank his parents, Syed Sirajul Islam and Kazi
Shahana Begum, for everything.


\bibliography{custom}

\appendix
\label{sec:appendix}

\section{Preliminaries: Token Attribution}
\label{sec:prelemTokenAttribution}
To identify which tokens in an input text sequence carry significant influence on the language model's decision-making process for a given task, we compute the token attribution scores using task-specific BERT-class encoders \cite{devlin2019bert}. We obtain the saliency scores from these models by adopting two state-of-the-art (SoTA) token attribution methods: the global token attribution analysis method (GlobEnc) \cite{modarressi-etal-2022-globenc} and the globally propagated locally decomposed attribution analysis method (DecompX) \cite{modarressi-etal-2023-decompx}, which are discussed in this section.



\subsection{GlobEnc}
GlobEnc aggregates the layerwise analysis method $\mathcal{N}_{\text{\textsc{Enc}}} \coloneqq (\Vert \thicktilde{x}_{i \leftarrow j} \Vert) \in \mathbb{R}^{n\times n}$ from input token $j$ to output token $i$, across all the encoder layers using attention rollout \cite{abnar-zuidema-2020-quantifying}. As elucidated in \citet{kobayashi-etal-2020-attention}, $\tilde{x}_{i\leftarrow j}$ approximates the attributed vector contribution from input token $j$ to output token $i$, computed as the weighted and transformed value vector $\alpha_{ij} f(x_j)$, where $f$ is the affine transformation to value space. The vector is adjusted for residual connections and layer normalization to capture the true \textit{mixed} influence beyond attention weights.
For each encoder layer, the output is computed as:
\begin{align}
    \label{eq:enc_output}
    \thicktilde{z}_{i \leftarrow j} &= g_{z_i^+}\left( \sum_{h=1}^H \alpha_{ij}^hf^h({x_j}) + \mathbbm{1}[i=j]x_i \right)\\
    \thicktilde{x}_{i \leftarrow j} &\approx g_{\thicktilde{z}_i^+}(\thicktilde{z}_{i \leftarrow j}) = \frac{\thicktilde{z}_{i \leftarrow j}-\operatorname{mean}(\thicktilde{z}_{i \leftarrow j})}{\operatorname{std-dev}(\thicktilde{z}_{i}^+)}\odot\gamma
\end{align}
such that $\mathcal{N}_{\text{\textsc{ResLN}}} \coloneqq (\Vert \thicktilde{z}_{i \leftarrow j} \Vert) \in \mathbb{R}^{n\times n}$ is the attribution analysis to measure the influence of an encoder layer's input token $j$ on its corresponding output token $i$, considering the residual connections and layer normalizations in the attention block. Here, $\alpha_{ij}^h$ denotes the raw attention weight from the $i$th token to the $j$th token in the attention head $h \in \{1, \dots, H\}$, collectively forming the set of attention weights $\mathbf{A}^h = \{\alpha_{ij}^h \bigm| 1\leq i,j\leq n\}$ for the $h$th attention head and $f^h(x)$ is the transformation function as defined in \citet{kobayashi-etal-2020-attention}. For layer $l$, the attention rollout with respect to the input is computed as per the following recursive function:
\begin{align}
    \label{eq:att_rollout}
    \thicktilde{\mathbf{A}}_l &= 
    \begin{cases}
        \hat{\mathbf{A}}_l \thicktilde{\mathbf{A}}_{l-1} \quad\text{; if}\quad l > 1\\
        \hat{\mathbf{A}}_l \quad\quad\quad\, \text{; if}\quad l = 1
    \end{cases}\\
    \hat{\mathbf{A}}_l &= 0.5\bar{\mathbf{A}}_l + 0.5\mathbf{I}
\end{align}
where $\bar{\mathbf{A}}_l$ is the mean raw attention grid across all the heads relevant to the layer $l$.

\subsection{DecompX}

While GlobEnc aggregates layer-wise analyses with a modified attention rollout and approximates FFNs via residuals and scalar norms, it risks information loss. DecompX improves this by propagating the locally decomposed vectors throughout the layers to construct a global decomposition, including FFNs with linear activation approximations, and extending attributions to prediction-based values. Formally, the locally decomposed vector at layer $\ell$ is expressed as:
\begin{align}
    \thicktilde{z}^{\ell}_{i} &= \mathrm{LN}\left( \sum_{k=1}^{N} \left[ \mathbf{x}^{\ell}_{i \Leftarrow k} + \mathbf{z}^{\ell}_{i \Leftarrow k} \right] \right),
\end{align}

where the normalized contribution is given by:

\begin{equation}
    \begin{aligned}
        \mathrm{LN}\left(\mathbf{z}_i^{+ \ell}\right) &= \sum_{k=1}^{N} \underbrace{g_{z^{+\ell}_i} \left( \mathbf{z}_{i \Leftarrow k}^{+ \ell} \right)}_{\tilde{\mathbf{z}}_{i \Leftarrow k}^{\ell}} + \boldsymbol{\beta} \\
g_{z^{+\ell}_i} \left( \mathbf{z}_{i \Leftarrow k}^{+ \ell} \right) &:= \frac{\mathbf{z}_{i \Leftarrow k}^{+ \ell} - m\left(\mathbf{z}_{i \Leftarrow k}^{+ \ell} \right)}{s\left(\mathbf{z}_i^{+ \ell} \right)} \odot \boldsymbol{\gamma}
    \end{aligned}
\end{equation}
DecompX explicitly integrates all encoder layer components, notably the non-linear feed-forward networks, to enhance representational fidelity. The $\text{FFN}$ operation is defined as:
\begin{equation}
\begin{aligned}
\mathbf{z}_{\text{FFN}}^{\ell} &= \text{FFN}(\tilde{\mathbf{z}}_i^{\ell}) \\
&= f_{\text{act}}( \underbrace{\tilde{\mathbf{z}}_i^{\ell} \mathbf{W}_{\text{FFN}}^{1} + \mathbf{b}_{\text{FFN}}^{1}}_{\bm{\zeta}_i^{\ell}}) \mathbf{W}_{\text{FFN}}^{2} + \mathbf{b}_{\text{FFN}}^{2}
\end{aligned}
\end{equation}
\begin{equation}
    \begin{aligned}
        z_{\text{FFN},i}^{\ell} &= f_{\text{act}}^{(\zeta_i^{\ell})} \left( \sum_{k=1}^{N} \zeta_{i \Leftarrow k}^{\ell} \right) W_{\text{FFN}}^2 + b_{\text{FFN}}^2 \\
&= \sum_{k=1}^{N} \underbrace{ \theta(\zeta_i^{\ell}) \odot \zeta_{i \Leftarrow k}^{\ell} + b_{\text{FFN}}^2 }_{z_{\text{FFN}, i \Leftarrow k}^{\ell}}
    \end{aligned}
\end{equation}

A propagation mechanism is implemented for decomposed token representations across multiple layers, preserving their distinct identities and preventing representation conflation:
\begin{equation}
    \begin{aligned}
        x_i^{\ell+1} &= \operatorname{LN} \left( \sum_{k=1}^{N} \underbrace{ \left[ \tilde{z}_{i \Leftarrow k}^{\ell} + z_{\mathrm{FFN}, i \Leftarrow k}^{\ell} \right] }_{z_{\mathrm{FFN}+, i \Leftarrow k}^{\ell}} \right) \\
&= \sum_{k=1}^{N} \underbrace{ {g_{z_{\mathrm{FFN}+,i}^{\ell}}} \left( z_{\mathrm{FFN}+, i \Leftarrow k}^{\ell} \right) + \beta }_{x_{i \Leftarrow k}^{\ell+1}}
    \end{aligned}
\end{equation}
The final classification mechanism uses these decomposed vectors to quantitatively assess the precise directional impact (positive or negative) of each input token on specific target classes.
The established formulations within this section permit the derivation of $x_{i \Leftarrow k}^{\ell+1}$ from its antecedent, $x_{i \Leftarrow k}^{\ell}$. Through the iterative application of this procedure across the entirety of the layered architecture, the final value, $x_{i \Leftarrow k}^{L+1}$, is consequently ascertained. For the task-specific classification head,
\begin{equation}
\begin{aligned}
y_c = \sum_{k=1}^{N} y_{c \Leftarrow k}
\end{aligned}
\end{equation}
where the term $y_{c \Leftarrow k}$ quantifies the specific contribution of the $k{\text{th}}$ token towards the overall prediction score.

\section{Proof}
\label{sec:proof}
The proof proceeds in three steps: (i) order the deletions arbitrarily and telescope $f(D)$ into a sum of marginals; (ii) reduce each marginal $\Delta_{d_r} f(A_{r-1})$ to a singleton marginal $\Delta_{d_r} f(\emptyset)$ by peeling tokens off $A_{r-1}$ one at a time, accumulating at most $\gamma_\tau(r-1)$ slack via Assumption~\ref{assump:pairwiseBound}; and (iii) bound each singleton marginal by $C_\tau s_{d_r}$ via Assumption~\ref{assump:del}. The bound is symmetric in $D$, so the choice of ordering $d_1,\dots,d_q$ is immaterial.
\begin{proof}
We order $D=\{d_1,\dots,d_q\}$ and let $A_r:=\{d_1,\dots,d_r\}$ be the set of tokens deleted after $r$ steps and $A_0=\emptyset$. 
For each $r$ and $u\in A_{r-1}$,
\begin{align*} 
&\Delta_{d_r,u} f(A_{r-1}\setminus\{u\})\le\gamma_\tau \tag*{[\text{Eq.} \ref{eq:bounded-interaction}]}
\\\implies&\Delta_{d_r}\Delta_u f(A_{r-1}\setminus\{u\})
\;\le\; \gamma_\tau\tag*{[\text{Eq.} \ref{eq:second-def}]} \\
\implies&\Delta_{d_r} f(A_{r-1})-\Delta_{d_r} f(A_{r-1}\setminus\{u\}) \le {\gamma_\tau}\\
\implies&\Delta_{d_r} f(A_{r-1}) \le \Delta_{d_r} f(A_{r-1}\setminus\{u\}) + \gamma_\tau.
\end{align*}

Iterating over all $u\in A_{r-1}$ gives
\begin{align*}
    \Delta_{d_r} f(A_{r-1}) &\le \Delta_{d_r} f(\emptyset) + \gamma_\tau(r-1)\\ 
    &{=} f(\{d_r\})-f(\emptyset)+\gamma_\tau(r-1) \\ 
    &{=} f(\{d_r\}) + \gamma_\tau(r-1)\\ 
    &\le C_\tau s_{d_r} + \gamma_\tau(r-1).\quad\quad [\textup{Eq. \ref{eq:singleton-dom}}]
\end{align*}
Formulating the telescopic sum: 
\begin{align*}
    f(D)=f(A_q)&=\sum_{r=1}^q(f(A_r)-f(A_{r-1}))\\
    &=\sum_{r=1}^q \Delta_{d_r} f(A_{r-1}) \tag*{[\text{Eq.} \ref{eq:firstDef}]}\\
    &\le \sum_{r=1}^q C_\tau s_{d_r} + \gamma_\tau \sum_{r=1}^q (r-1).
\\\text{Hence, } f(D)&\le C_\tau\sum_{i\in D} s_i + \frac{\gamma_\tau}{2}q(q-1).~~~~~~\qedhere
\end{align*}
\end{proof}
\section{On Handling Long Inputs}
\label{app:longInputs}
The token-attribution backbones used by {\shortName{}} are BERT/RoBERTa-style
encoders, whose self-attention cost and learned positional embeddings impose a
fixed maximum input length (512 subword tokens in the models used here). Rather
than feeding the entire prompt or document to the attribution encoder at once,
we {use} a sentence-level chunking strategy. Given an input document
$x = (w_1,\ldots,w_n)$, we first segment it into sentences
$x = \{s_1,\ldots,s_m\}$ using a sentence tokenizer. Each sentence $s_i$ is then
tokenized independently with the corresponding BERT/RoBERTa tokenizer and passed
through the attribution model as a separate encoder input. This ensures that the
encoder never receives the full long document as a single sequence.

For DecompX, each sentence-level input is encoded with tokenizer-level
truncation enabled, so any pathological sentence that exceeds the encoder limit
is clipped to the model's maximum sequence length before attribution is
computed. DecompX then decomposes the classifier output over the subword tokens
of that sentence. Let $z_{i,j}$ denote the $j$-th subword token in sentence
$s_i$ and let $a^{\mathrm{Dec}}_{i,j}$ be its DecompX attribution score. The
scores are min--max normalized within the sentence-level forward pass:
\[
\tilde{a}^{\mathrm{Dec}}_{i,j}
= \frac{a^{\mathrm{Dec}}_{i,j} - \min_k a^{\mathrm{Dec}}_{i,k}}
       {\max_k a^{\mathrm{Dec}}_{i,k} - \min_k a^{\mathrm{Dec}}_{i,k} + \epsilon},
\]
{where $\epsilon$ is a small constant for numerical stability.}

For GlobEnc, each sentence is likewise processed independently. The model
returns layer-wise attention/norm information for the sentence-level input, and
GlobEnc computes the rollout flow from the classification token to the input
subwords. If $R_i$ is the rollout matrix for sentence $s_i$, the token score is
computed from the classification-token row of the flipped layer-wise rollout and
averaged across layers:
\[
\tilde{a}^{\mathrm{Glob}}_{i,j}
= \frac{1}{L}\sum_{\ell=1}^{L}
    \frac{R^{(\ell)}_{i,\mathrm{[CLS]},j}}
         {\max_k R^{(\ell)}_{i,\mathrm{[CLS]},k} + \epsilon}.
\]

After sentence-level attribution, the token--score pairs from all sentences are
concatenated back in their original document order. The frugalized prompt is
formed by ranking tokens globally by their attribution scores, retaining the top
$k\%$ of tokens, and restoring the retained tokens to their original order.
Optionally, a sentence-importance weight $\lambda_i$ (\textit{e.g.}, LexRank) can be
applied before aggregation, yielding
$\hat{a}_{i,j}=\lambda_i\tilde{a}_{i,j}$, although the core long-context
handling remains the same. This procedure avoids document-level truncation while
preserving local syntactic context for attribution and permitting arbitrary-length
prompts to be scored by fixed-window BERT/RoBERTa encoders.
\section{Experimental Details}
\label{sec:app_experimentalDetails}

\subsection{Tasks}
\label{subsec:tasks}
We evaluate the efficacy of the token reduction by experimenting across four NLP tasks: sentiment analysis (text classification), summarization (text generation), commonsense question answering, and mathematical reasoning. These tasks span both discriminative and generative paradigms, enabling evaluation of the LLM's linguistic and cognitive demands under diverse conditions. 

\subsection{Dataset}
\label{subsec:dataset}

For sentiment analysis, we use {the} IMDb movie review {dataset} \cite{maas-etal-2011-learning-imdb}, a binary sentiment analysis corpus comprising movie reviews labeled positive or negative. We use Argilla News to evaluate abstractive summarization on news articles \cite{ahmed2017detectionArgillaNewSum}. For commonsense QA, we use CosmosQA \cite{huang-etal-2019-cosmos}, which primarily focuses on commonsensical and causal reasoning within short narrative passages. The questions stipulated in CosmosQA are easy for humans to solve, but usually challenging for LLMs. We use GSM8k \cite{cobbe2021trainingMathReasoning} for mathematical reasoning, which contains grade-school {math word problems (MWPs)} to assess basic arithmetic reasoning and multi-step logical deduction. For evaluating our method, we take $1000$ random samples from each dataset.

\subsection{{Models}}
\label{subsec:models}

{We perform inference using a set of contemporary frontier models, varying in architecture and scale, to examine the generalizability of our method. The models are: Llama-3 8B, Llama-3 70B \cite{meta2024llama3}, GPT-3.5 \cite{brown2020language}, Gemini 2.0 Flash Thinking \cite{googledm2025gemini25}, and o3-mini \cite{openai2025o3mini}. The closed-source models are accessed via their respective APIs, and the open-source models are accessed via OpenRouter\footnote{\label{fn:openrouter}\url{https://openrouter.ai/}}, using standardized prompts to minimize formatting-induced variability.}

\subsection{{Evaluation Metrics}}
\label{subsec:evaluationMetrics}
{We employ task-specific metrics to evaluate performance across both full and frugalized setups. For sentiment classification, we report accuracy and F1 score. Summarization performance is assessed using BLEU, ROUGE-1, ROUGE-2, ROUGE-L, BERTScore, and METEOR to capture both superficial and semantic quality of the responses. For commonsense question answering, we use accuracy as the dataset follows a multiple-choice format. Mathematical reasoning is evaluated using pass@$1$, which measures the proportion of problems correctly solved on the first attempt.}

\end{document}